# Social Media Personal Event Notifier Using NLP and Machine Learning

Pavithiran Ganeshkumar
*Department of Computer Science and Engineering*
*Panimalar Institute Of Technology*
Chennai, Tamil Nadu, India
pavithiran.nasa@gmail.com

Sharan Padmanabhan
*Department of Computer Science and Engineering*
*Panimalar Institute Of Technology*
Chennai, Tamil Nadu, India
sharanmozart@gmail.com

Ashwin Kumar BR
*Department of Information Technology*
*Panimalar Institute Of Technology*
Chennai, Tamil Nadu, India
ashwinkumarbr27@gmail.com

Vetriselvi A
*Department of Computer Science and Engineering*
*Panimalar Institute Of Technology*
Chennai, Tamil Nadu, India
vetriselviayyachami@gmail.com

*Abstract*—Social media apps have become very promising and omnipresent in daily life. Most social media apps are used to deliver vital information to those nearby and far away. As our lives become more hectic, many of us strive to limit our usage of social media apps because they are too addictive, and the majority of us have gotten preoccupied with our daily lives. Because of this, we frequently overlook crucial information, such as invitations to weddings, interviews, birthday parties, etc., or find ourselves unable to attend the event. In most cases, this happens because users are more likely to discover the invitation or information only before the event, giving them little time to prepare. To solve this issue, in this study, we created a system that will collect social media chat and filter it using Natural Language Processing (NLP) methods like Tokenization, Stop Words Removal, Lemmatization, Segmentation, and Named Entity Recognition (NER). Also, Machine Learning Algorithms such as K-Nearest Neighbor (KNN) Algorithm are implemented to prioritize the received invitation and to sort the level of priority. Finally, a customized notification will be delivered to the users where they acknowledge the upcoming event. So, the chances of missing the event are less or can be planned.

## I. INTRODUCTION

Communication entails more than mere transmission of information. It also entails communicating messages and critical events in various ways, such as verbal communication, nonverbal communication, or visualizations [1]. Nowadays, people communicate more on mobile devices through social media platforms. Social media helps people connect, socialize, and stay informed about current events in an age where technology is dominant. Today's users (people) use social media to communicate their feelings, organize events, form meaningful connections with others, know about significant events, and stay informed (what is happening in daily life). Despite their love for social media, many users struggle to find the time to use these platforms.

As a result, people miss important events and meetings that others have posted and arranged, which influences their personal or professional lives.

In general, we see tons of messages such as news, schedule dates, simple chat communications, or business advertisement messages that are constantly popping up on the notification bar. They urge people and exasperate them to check the overall messages alone and sometimes ignore them. Thus, people tend to lose important dates, which are memorable and important meetings or schedules. Providing preference to specific messages helps the user to get well informed about important future events. It even maintains a live update on previous events and meetings.

Meetings and events are important aspects of social and commercial life. Meetings and events, whether they be personal or professional, frequently determine a person's destiny. Given its significance, this must be treated seriously by emphasizing specific messages when communicating [2].

This study focuses on the solution to the aforementioned problem by creating an autonomous system that can notify the user of critical event changes [3]. The data is first obtained from the social media application (scraping social media data). Natural Language Processing (NLP) techniques are then applied to analyze the collected information from the text file and identify all of its important properties. These results are examined by the machine learning model for event prioritization [4]. Finally, based on the prioritized information, the user will receive a custom notification that will be displayed and pinned until the user acknowledges it.

## II. LITERATURE SURVEY

Over the past 20 years, information and communication technology has undergone a fast transformation, with the development of social media being a huge step forward. The rate of change is quickening. For instance, the development of mobile technology has a tremendous impact on social media influence. Globally, mobile devices account for the vast majority of internet time. They give everyone access to the resources needed to connect using any device, from any location, at any time [5].

Hundreds of billions of people post pictures and videos of anything from crowded events to their closest family moments. Social media platforms allow us to interact with friends and establish new relationships like never before, yet these improvements in communication and social connection might not be without a price. Researchers found that people who documented and shared their experiences on social media created fewer accurate recollections of such events, according to a recent study published in the Journal of Experimental Social Psychology [6]. Also, the busy schedule of individuals makes them forget important events.

A deep learning system along with Regular Expression and NLP can assist in finding critical messages by categorizing them and notifying the individual.

A regular expression, consisting of a string of characters, defines a querying pattern that is typically used by string searching algorithms for input validation or "find" or "find and replace" actions on strings. Techniques for regular expressions are developed in theoretical computer science and formal language theory. Search engines, word processors, text editors, text processing tools like sed and AWK, and lexical analysis use regular expressions [7].

Natural Language Processing (NLP) is a subfield of semantics, software engineering, and artificial intelligence dealing with the coordination between computers and Diverse human languages [8]. Their concept was an efficient method for automatic text summarization by using Natural Language Processing (NLP) and Machine Learning (ML) techniques. Like opening up a web browser and searching for specific data in the search box [9]. The web browser shows the result, the result which will be summarized by using understandable python programming. The end result of the program shows the final copy of the summarized version of the topic in which they have been searching for relevant questions. Thus, the NLP comes with text classification and processing.

## III. DESIGN AND METHODOLOGY

In this paper, the text is scraped in order to collect data. The data is processed using steps like text pre-processing and name-entity recognition. Events are prioritized using the machine learning model to determine the level of priority after message analysis. Finally, a customized notification will be sent to the user notifying the event.

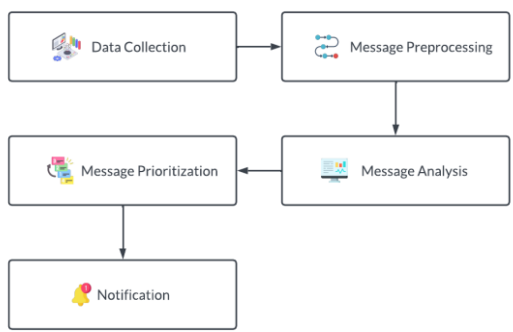

Fig. 1 – Phases of Methodology

### A. Data Collection:

The data is gathered through social media apps like WhatsApp, Telegram, etc., and is typically in the form of a text file which is accomplished using the Python Selenium Library [10]. This selenium tool will automatically enter the web portal of social media and download or scrape the conversation data from there. It will be used to browse across the web and access the necessary social media platform that supports the web. The data will then be transmitted for text processing after being scraped.

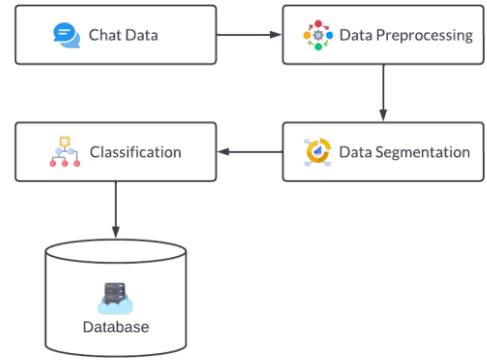

Fig. 2 – Data Collection Process

As shown in Figure 2, the process begins with a chat application, where the data of a particular user or a group is taken into consideration and the conversation data is analyzed. Data segregation is then carried out in order to categorize the data according to the entities. Finally, the storage is used to store the conversation data.

### B. Message Preprocessing and Analysis:

After the messages are collected from social media, the resultant text file will be processed by using Natural Language Processing which will filter the unnecessary information from the received message [11].

- Text Preprocessing:

Text preprocessing is done to clean and transform the collected messages. After the preprocessing of the messages, the data is ready to be given to the Machine Learning model [12].

```
--- original ---
I am inviting you my brother's wedding which is on 1 august.
The reception starts at 6pm and the marriage starts at 10 in the morning.
--- cleaning ---
i am inviting you my brother's wedding which is on 1 august
The reception starts at 6pm and the marriage starts at 10 in the morning
--- tokenization ---
['i', 'am', 'inviting', 'you', 'my', 'brother', 's', 'wedding', 'which',
'is', 'on', '1', 'august']
['the', 'reception', 'starts', 'at', '6', 'pm', 'and', 'the', 'marriage',
'starts', 'at', '10', 'in', 'the', 'morning']
```

Fig. 3 – Data Collection Process

- Tokenization:

Tokenization is a step in the NLP process that divides large strings of text into smaller parts, or tokens. Larger pieces of text can be tokenized into phrases, and sentences can be tokenized into words, and so on. After a piece of text has been appropriately tokenized, additional processing is typically conducted. This will make it easier to comprehend collected data or to analyze data to decipher its significance. For instance, as shown in Fig. 3, the chat data is tokenized, or broken down into words and phrases [13].

- Stop Words Removal:

In the majority of languages, the most frequent words are articles, prepositions, pronouns, conjunctions, etc. These words offer little information to the text, thus we will remove them to filter the data. So, after the words have been tokenized, stop words such as "the," "is," "was," "in," and so on will be eliminated. By deleting these words, it makes text processing in NLP more viable [14].

- Named Entity Recognition (NER):

After the stop words have been eliminated, the NER will be utilized to locate and classify important textual content. Any word or set of words that consistently refers to the same item is referred to as an entity. Each newly discovered entity is divided into one of many groups.

As seen in the aforementioned Fig.3. The entity will be divided into many categories inside that text, such as person, place, time, organization, events, etc. In order to find and classify the relevant information in the provided text data, we will be creating our own unique entity in this case [15].

- Regex Library (Regular Expression):

A regex is an array of words that creates patterns that assist in matching, locating, and managing texts. It identifies the date and time from the preprocessed text and creates an event using a Machine learning model. The data which is identified by the regular expression will be stored in a text or JSON file.

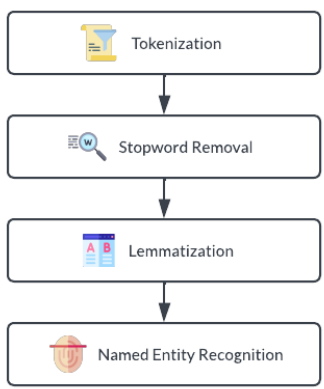

Fig. 4 – Processing Of Data

As a result, the stored data is processed in accordance with the criteria, as shown in Figure 4. Date, time, and event data are taken into consideration as of now. Regex, also known as the regular expression, is used to identify these data and separates them and name entity recognition is used to obtain the event data. Once the data collection processes are complete, the deep learning model will be utilized to develop and deliver a personalized notification. The iterated database will contain the modified data [16].

- Lemmatization:

Lemmatization is the process of reducing provided words to root words. As can be seen, the lemmatization algorithm will be aware of the word's definition. When the term is reduced to its base word, it will know what the word actually means. It will produce accurate results since it utilizes a dictionary to condense the words. And without understanding the definition of the word, it won't alter it. The term drove or driven and drives is lemmatized as drive, which is the root word [17].

C. Training the Model:

The processed texts are iterated by the deep learning model to prioritize the notification. This is accomplished by implementing a technique like the k-nearest algorithm to sort the unstructured data. The model will be trained using a predefined value based on the priority level and will prioritize the event. User preferences, such as varying schedules as well as needs for various persons, will be taken into account while creating training data. In order to ensure that users are notified of events with the highest reliability, the process of training data to notify the priority will be based on user desire or necessity. Additionally, prior user choices are taken into account while stacking the collected data. Each time a user updates their selections, the dataset or database is updated in accordance with the necessary patterns. As this is done, iterated data is used to train the model, which keeps it updated [18].

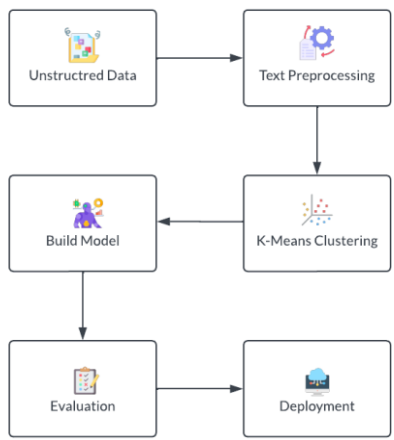

Fig. 5 – Deep Learning Process

As shown in Figure 5, the data present in iterated data in the database is then preprocessed and is processed through a Dense Embedding filter where it will segregate the data and is then iterated into a hidden layer, after processing the data is segregated according to events.

IV. RESULTS AND TEST CASES

To check the functionality, we made a test scenario with 15 participants. We conducted this test and created an analysis based on the daily phone usage of 15 social media users, and it has been found that the majority of them do not use their phones regularly. Remarkably, by performing the test, the system was able to notify and save the day for nearly 8 out of 15 users who were likely to miss a last-minute event that they've been invited through chat.

As a result, they were made aware of the event and were able to make arrangements in accordance with it, aiding in the creation of a well-organized timetable for any important events.

This model thus will provide a precise output. By training the model using deep learning, 70% of the training data set and 30% of the test data set has provided an accuracy and viable output and there is a possibility of high accuracy rate to be attained. Additionally, the data will be scraped and initialized in the first iteration before processing and filtered in the second iteration using NLP. The data will then be prioritized using machine learning models in the third iteration, which allows the delivery of a user-specific notice. Finally, it is notified to the consumers.

Since this model provides accurate event notification, derived from social media chat with the utmost care, anyone using this system may trust and hope the output as discussed in the outcome.

## V. CONCLUSIONS AND FUTURE WORK

The planned system will provide the solution to the fundamental problem of individuals who miss their critical communication and thus helps the users to receive time and date-sensitive notifications of events immediately.

This model will employ machine learning and NLP models to deliver the message to the users in a personalized manner by classifying the occurrences in accordance with how it has been taught. As a result, the weak social media users will also be immediately informed, which will be very helpful to individuals in their daily planning as well as scheduling events where they could be organized and prepared for any kind of future events or invitations.

The future enhancement will include a few extra potential event data kinds in order to provide customers with tailored notifications. Additionally, the priority will be determined based on the user's interactions with his contacts over time as well as their relationship to the user. By looking up and computing the user's phone-based timelines, the notification of the event to the user will also be improved. In order to make the system more interactive with the user. This will also soon see performance and speed enhancements.